\documentclass[sigconf,screen]{acmart}

\makeatletter
\def\@ACM@checkaffil{
    \if@ACM@instpresent\else
    \ClassWarningNoLine{\@classname}{No institution present for an affiliation}%
    \fi
    \if@ACM@citypresent\else
    \ClassWarningNoLine{\@classname}{No city present for an affiliation}%
    \fi
    \if@ACM@countrypresent\else
        \ClassWarningNoLine{\@classname}{No country present for an affiliation}%
    \fi
}
\copyrightyear{2023}
\acmYear{2023}
\setcopyright{acmlicensed}\acmConference[ICCPS '23]{ACM/IEEE 14th International Conference on Cyber-Physical Systems (with CPS-IoT Week 2023)}{May 9--12, 2023}{San Antonio, TX, USA}
\acmBooktitle{ACM/IEEE 14th International Conference on Cyber-Physical Systems (with CPS-IoT Week 2023) (ICCPS '23), May 9--12, 2023, San Antonio, TX, USA}
\acmPrice{15.00}
\acmDOI{10.1145/3576841.3585930}
\acmISBN{979-8-4007-0036-1/23/05}

\usepackage[utf8]{inputenc}
\usepackage{enumitem}
\setlist[enumerate]{label*=\arabic*.}
\usepackage{graphicx}
\usepackage{caption}
\usepackage{subcaption}
\usepackage{algorithm,algorithmic}
\usepackage{amsmath}
\usepackage{xfrac}
\usepackage{hyperref}
\usepackage{array}
\usepackage{booktabs}
\usepackage{mathtools}
\usepackage{forest}

\definecolor{mygreenii}{RGB}{91, 198, 98}
\definecolor{mygreeni}{RGB}{148, 188, 151}

\usepackage{titlesec}
\titlespacing{\section}{0pt}{12pt plus 4pt minus 2pt}{4pt}
\titlespacing{\subsection}{4pt}{6pt plus 2pt minus 2pt}{4pt}
\titlespacing{\subsubsection}{8pt}{6pt plus 4pt minus 2pt}{4pt}
\titlespacing{\paragraph}{12pt}{2pt plus 2pt minus 0pt}{2pt}
\newcommand{\tworowsubtablecenter}[2]{\begin{tabular}{@{}c@{}} #1 \\ #2 \end{tabular}}
\newcommand{\tworowsubtableleft}[2]{\begin{tabular}{@{}l@{}} #1 \\ #2 \end{tabular}}
\newcommand{\urlNewWindow}[1]{\href[pdfnewwindow=true]{#1}{\nolinkurl{#1}}}

\title{\texttt{AVstack}: An Open-Source, Reconfigurable Platform for Autonomous Vehicle Development}

\author{R. Spencer Hallyburton}
\affiliation{%
    \institution{Duke University}
    \city{Durham}
    \state{North Carolina}
}
\email{spencer.hallyburton@duke.edu}
\author{Shucheng Zhang}
\affiliation{%
    \institution{Duke University}
    \city{Durham}
    \state{North Carolina}
}
\email{shucheng.zhang@duke.edu}
\author{Miroslav Pajic}
\affiliation{%
    \institution{Duke University}
    \city{Durham}
    \state{North Carolina}
}
\email{miroslav.pajic@duke.edu}

\begin{document}


\begin{abstract}
Pioneers of autonomous vehicles (AVs) promised to revolutionize the driving experience and driving safety. However, milestones in AVs have materialized slower than forecast. Culprits include (1)~the lack of verifiability of proposed state-of-the-art AV components, and (2)~stagnation of pursuing next-level evaluations, e.g.,~vehicle-to-infrastructure (V2I) and multi-agent collaboration. In part, progress has been hampered by: the large volume of software in AVs, the multiple disparate conventions, the difficulty of testing across datasets and simulators, and the inflexibility of state-of-the-art AV components. To address these challenges, we present \texttt{AVstack}\footnote{\url{https://avstack.org/research}~\cite{avstack}.},\footnote{\url{https://github.com/avstack-lab}.}, an open-source, reconfigurable software platform for AV design, implementation, test, and analysis. \texttt{AVstack} solves the validation problem by enabling first-of-a-kind trade studies on datasets and physics-based simulators. \texttt{AVstack} 
addresses the stagnation problem as a reconfigurable AV platform built on dozens of open-source AV components in a high-level programming language. We demonstrate the power of \texttt{AVstack} through longitudinal testing across multiple benchmark datasets and V2I-collaboration case studies that explore trade-offs of designing multi-sensor, multi-agent algorithms.
\end{abstract}
\maketitle

\section{Introduction}

The AV industry has proliferated over the past two decades. Experts point to the DARPA Grand Challenge as the coming of age of AVs~\cite{2006darpagrand}. Soon after, expectations ballooned that we would see fully automated vehicles on the road within a decade~\cite{2017brownsocial}. In response, the autonomy community has exploded into industry and academic players both large and small. 
%
However, milestones in AV development have slowed in recent years. In fact, Tesla has promised to deliver fully self-driving cars ``next year'' for the last 8 years~\cite{2022teslapromise} and is as of yet still deploying Level-2 solutions. 

A major challenge to AV development is that most AV solutions are proprietary and closed-source. This is a result of the immense cost of development that safety-critical AVs require. However, the rush to deploy autonomous vehicles and the proprietary nature of industry solutions are conflicting. In particular, industry progress is outpacing research and development. This disparity is negatively impacting progress and trust in AVs. While industry rushes to capture a new market, access to representative platforms is hampering fundamental safety and performance research~\cite{2017cummingsreview}.

We find two culprits for such a slowdown in AV research. First, proposed state-of-the-art AV algorithms and components perform insufficient transfer testing and longitudinal analysis. This leads to a lack of accountability and verifiability. Second, much AV research pursues (marginal) improvements on single-component benchmarks (e.g.,~LiDAR-based detection challenges~\cite{2013kittidataset, 2020nuscenesdataset, 2020waymodataset}). Similarly, pursuing critical next-level evaluations such as multi-agent (e.g.,~vehicle-to-vehicle (V2V), vehicle-to-infrastructure (V2I)) collaborative sensing or safety \& security analysis has~stagnated.

At the root of these problems are several barriers: (1)~Designing and implementing an AV require large amounts of complex software. The jumps from testing components on static datasets to longitudinal datasets to full-stack simulations are large, and no existing platform can handle all scenarios. (2)~Data sources use different conventions for coordinates, reference frames, metrics, calibrations, and more, which makes case-by-case conversions prone to error. (3)~Mature AV platforms are designed with custom messaging protocols in low-level languages, which creates rigid AV architectures and implementations. Rigidity inhibits modular testing and puts a high-barrier on design changes. (4) Implementations of state of the art AV algorithms are highly tailored towards benchmark challenges. Adapting them to new contexts is time-consuming.

Several platforms have emerged to support open-source AV development. Baidu's Apollo~\cite{BaiduApollo} and Autoware~\cite{Autoware} are established as preeminent platforms for deployable, real-time AVs. Recently, Pylot~\cite{2021pylot} has also allowed for more trade studies in AVs (i.e.,~evaluating the impact of parameter/configuration changes on output metrics). Each of these platforms is useful and needed. However, each has serious shortcomings when it comes to the design of novel AV architectures for next-level challenges and transfer testing of AV components between datasets/simulators.

To fill this void, we present \texttt{AVstack}, a new research platform for the design, implementation, test, and analysis (DITA) of AVs. \texttt{AVstack} has the following four key innovations designed to promote modular AV design, simple implementation, wide-reaching testing, and insightful analysis. The key innovations~are:

\vspace{4pt}\noindent (1) \textit{Wide compatibility:} To our knowledge, \texttt{AVstack} is the first platform compatible with \emph{both benchmark AV datasets and physics-based AV simulators}. It maintains compatibility with dozens of \emph{open-source AV algorithms} across AV components and established metrics.

\vspace{2pt}\noindent (2) \textit{Unified conventions:} \texttt{AVstack} implements a flexible set of coordinate conventions attached to all vector-type objects to unify coordinates. It also unites component-wise metrics from multiple providers. This helps maintain forward and backward compatibility, reduces the user burden to keep track of case-by-case uniqueness, and enables complex, multi-sensor, multi-agent~configurations.

\vspace{2pt}\noindent (3) \textit{Modular testing:} \texttt{AVstack} streamlines the DITA phases of an AV lifecycle. Reconfigurable architectures break rigid constraints of prior platforms allowing for novel AV designs and reusable implementations. With \texttt{AVstack}, testing is performed seamlessly on both static/longitudinal datasets and physics-based simulators with diverse metrics at every AV component.

\vspace{2pt}\noindent (4) \textit{Low barrier adoption:} \texttt{AVstack} is written in a high-level programming language to allow for rapid prototyping and reusable implementations. A suite of AVs can be designed, implemented, tested, and analyzed with little effort in unique configurations such as multi-sensor, multi-agent settings.

\vspace{2pt}
\texttt{AVstack} is a framework for AV development. It provides a unique combination of performance and modularity -- high-performing algorithms with a flexible and reconfigurable architecture to enable diverse and rapid prototyping. \texttt{AVstack} is not the ``best'' framework for all AV applications. However, its innovations fill important voids in validating state of the art results, transfer testing between datasets/simulators, standardizing AV evaluations, and pursuing next-level questions in multi-sensor, multi-agent configurations. \texttt{AVstack} is available open-source\footnote{\url{https://github.com/avstack-lab}.}.

In summary, \texttt{AVstack} contributes the following innovations:
\begin{itemize}
    \item Unifies testing and analysis within and between benchmark static/longitudinal datasets and physics-based simulators. 
    \item Enables reconfigurable and reusable AV design through standardized interfaces and open-source support.
    \item Unifies disparate coordinate conventions to achieve forward \textit{and} backward compatibility to data sources.
    \item Streamlines component-wise metrics in all cases ranging from single-component analysis on static datasets to full-stack AVs on longitudinal situations.
    \item Promotes easy transfer testing between datasets and simulators with low-barrier trade study configuration tables.
    \item Provides a standardized interface for training supervised learning models on datasets and the CARLA simulator~\cite{2017carla}.
    \item Facilitates testing multi-sensor, multi-agent scenarios.
\end{itemize}

\subsubsection*{Terminology.}
In this work, we use the following terminology:
\begin{itemize}
    \item \textit{Algorithm:} A specific implementation of an AV component; e.g.,~PointPillars~\cite{2019pointpillars} is an algorithm.
    \item \textit{Component:} A generalization and grouping over algorithms; e.g.,~PointPillars~\cite{2019pointpillars} and 3DSSD~\cite{20203dssd} fall under the ``3D object detection'' component. 
    \item \textit{Module:} A grouping of similar components under a goal; e.g.,~2D \& 3D object detection $\rightarrow$ ``perception'' module.
    \item \textit{Architecture:} A designed connection of \textit{components} that will process sensor data and output control signals or state.
    \item \textit{Implementation:} A connection of specific algorithms that defines one particular realization of an AV architecture. 
\end{itemize}

\subsubsection*{Organization.} Section~\ref{sec:2-related-work} summarizes related efforts and their shortcomings for AV DITA. Section~\ref{sec:3-design} expands on challenges to AV research and how key design decisions allow \texttt{AVstack} to overcome these obstacles. Section~\ref{sec:4-use-cases} provides use-cases in longitudinal and multi-agent sensing demonstrating that \texttt{AVstack} enables new AV DITA capability. We finish with concluding remarks in Section~\ref{sec:conclusion}.
\section{Related Work} \label{sec:2-related-work}

\paragraph{Deployable AV Systems.}
Baidu Apollo~\cite{BaiduApollo} and Autoware~\cite{Autoware}~are 
highly adopted and stable AV repositories. Each have an architecture philosophy and have provided specific implementations. Both are designed for self-driving and have established relationships with industry to deploy in physical systems. Apollo is built on the CyberRT message passing framework while Autoware uses ROS~\cite{2009ros}. While both achieve high levels of performance, both struggle to maintain accessible APIs for research-level development. Both have high learning curves, are difficult to modify, and require powerful computers. Thus, both are ill-suited to perform longitudinal testing and reconfigurable prototyping on important benchmarks.

\paragraph{AV Research Platforms.}
Pylot~\cite{2021pylot} is an AV architecture in the Python language. Sensor data is passed using the low-latency, low-copy ERDOS~\cite{2021pylot} framework. Pylot provides an accessible interface where developers can compare algorithms within established components on the CARLA simulator~\cite{2017carla}. Pylot demonstrated near-real-time capability on a real system at low speeds. While Pylot maintains an accessible API, it is limited to the CARLA simulator and real-world self-driving; it cannot be tested on benchmark datasets. While it supports different algorithms within each component, the component architecture is fixed. It is not suitable for next-level questions such as multi-agent, collaborative sensing nor does it support end-to-end learning-based implementations.

OpenAI maintains the \texttt{gym} framework for evaluating reinforcement learning in episodic tasks. Researchers have used \texttt{gym} to develop control algorithms in self-driving using the AirSim~\cite{2018airsim} and TORCS~\cite{2000torcs} simulators. \texttt{gym} is not well-suited for component-wise evaluations in AVs and is designed only as a tool for training reinforcement learning algorithms.

\paragraph{General Frameworks.}
ROS~\cite{2009ros} provides a communication infrastructure above operating systems for multi-component robotics applications. ROS handles message passing between peer nodes in full-stack robotics case studies. In this way, ROS has greatly streamlined the development process for deployable robotics and has been recognized as a major research platform.

ROS and \texttt{AVstack} serve different purposes and can be used in a complementary way. \texttt{AVstack} handles development environments for DITA in both single-component and multi-component settings. \texttt{AVstack} is more suitable for rapid prototyping of AV algorithms and components while ROS is designed to handle communication between components for cyber physical systems (CPS) over potentially heterogeneous networks. For a full stack-simulation or a physical implementation, the two can be complementary: ROS can managed message passing and computation resources while \texttt{AVstack} can provide components and analysis.

\section{AVstack Key Design Decisions} \label{sec:3-design}

\begin{figure*}[t!]
    \centering
    \includegraphics[width=0.962\linewidth]{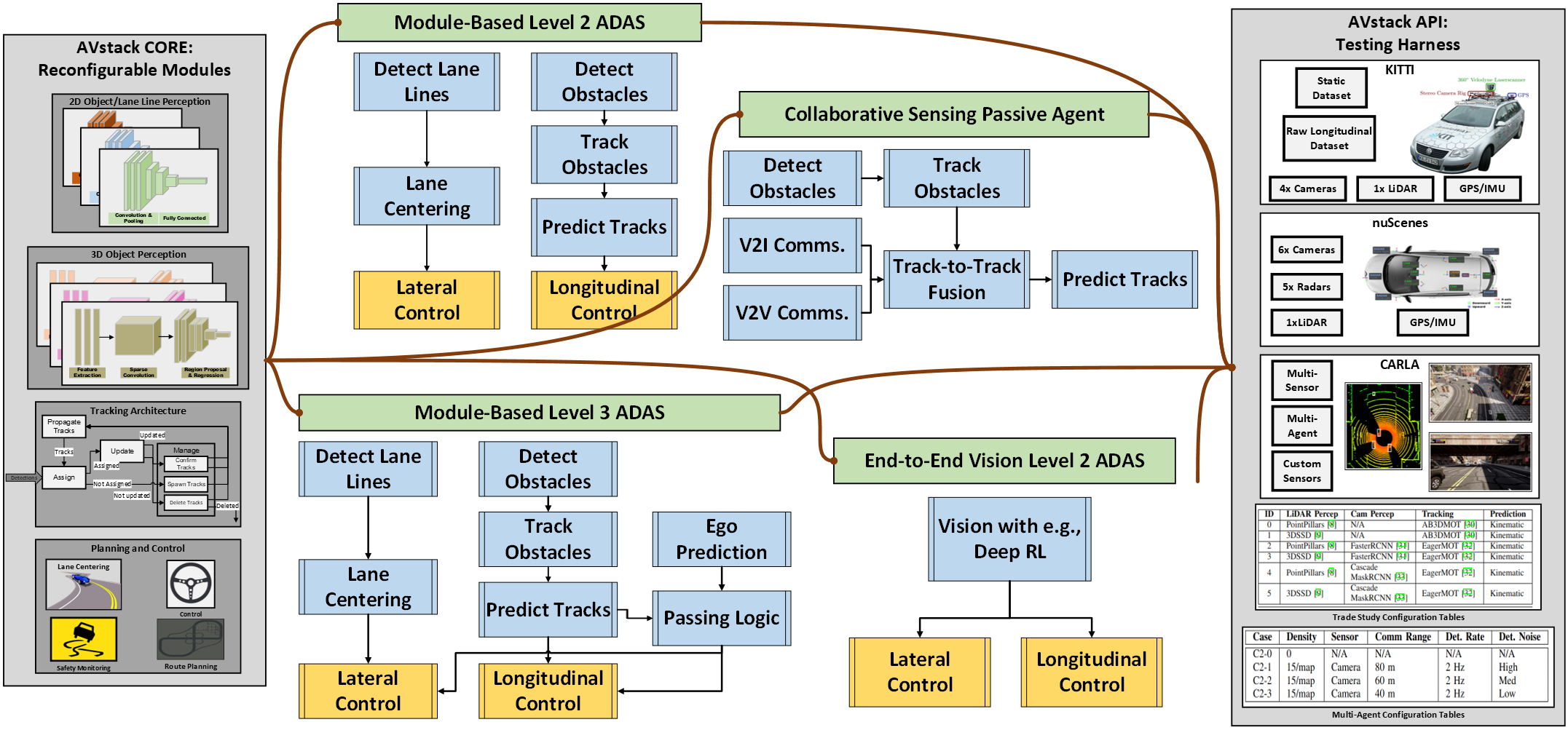}
    \vspace{-10pt}
    \caption{(left) \texttt{AVstack}'s core library provides source-agnostic modular components within a reconfigurable architecture. \texttt{AVstack} unifies diverse conventions from open-source providers to deliver community-vetted AV design. (center) With the core library, developer can specify many AV architectures, including ``active" (with control) and ``passive" (without control). Shown are a sample of architectures used in this work. (right) \texttt{AVstack}'s API library provides data source interfaces to the most popular of AV benchmarks and simulators. API also provides low-barrier trade-study capability with configuration tables.}
    \label{fig:avstack-full-diagram}
\end{figure*}

A recent slowdown in AV development is a consequence of at least two factors. First, proposed state-of-the-art AV algorithms and components perform insufficient trade studies, transfer testing, and longitudinal analysis. Second, a platform is needed that allows multi-component evaluations and lowers the barrier of pursuing next-level evaluations such as V2V and V2I collaboration. 

\texttt{AVstack} was designed to address the above shortcomings and more. In this section, we present the high-level innovations of \texttt{AVstack} that have allowed it to uniquely fill this large void in AV DITA. As illustrated in Fig.~\ref{fig:avstack-full-diagram}, \texttt{AVstack}'s key innovations have allowed for great strides in design modularity and robust testing~\&~analysis. The critical design decisions of \texttt{AVstack} fall under:
\begin{itemize}
    \item \textbf{Wide Compatibility:} \texttt{AVstack} is widely compatible with benchmark AV datasets and physics-based AV simulators. \texttt{AVstack} leverages many open-source AV components.
    \item \textbf{Unified Conventions:} \texttt{AVstack} standardizes coordinate conventions and maintains backward compatibility with legacy conventions. \texttt{AVstack} metrics and evaluations are expanded over many preceding benchmarks.
    \item \textbf{Modular Testing:} AVs can be designed quickly and flexibly in \texttt{AVstack} drawing from a bank of reconfigurable components. \texttt{AVstack} tests implementations seamlessly on static datasets, longitudinal datasets, and AV simulators with minimal software changes.
    \item \textbf{Low Barrier:} \texttt{AVstack} is written in a high-level programming language. A user can design Level 2-5 AVs and test in just dozens of lines of code. Trade studies can be initiated with simple configuration tables.
\end{itemize}

In the following, we present the motivations and high-level design details for \texttt{AVstack}. Each section begins by identifying specific barriers in AV development and how \texttt{AVstack} was designed as a solution to those challenges. We identify some intentional omissions from the design in Appendix~\ref{app:design-omissions}.

\subsection{Design Goal 1: Wide Compatibility} \label{sec:3-design-1-compatibility}

A community-supported foundation is of the utmost concern in \texttt{AVstack}. To ensure utility and staying power, \texttt{AVstack} was designed to be widely compatible with gold-standard benchmark datasets and simulators. To support representative AV design and implementation, \texttt{AVstack} maintains compatibility with many open source components. Below, we provide details on this compatibility.

\subsubsection{Design Goal 1.1: Inferfaces \& APIs} \label{sec:3-design-1.1-datasets_sims}
\paragraph{\textbf{Motivation.}}
The challenge to designing a widely-compatible platform is in supporting backward-compatibility and preparing for forward-compatibility without requiring constant overhauls that precipitate uncontrolled software bloat. This task is difficult enough that, until now, we had yet to see a platform bridge the dataset-simulator gap and deliver inter-source compatibility. In the following, we investigate several challenges.

\paragraph{Datasets.}
The KITTI dataset~\cite{2013kittidataset} changed the world of autonomous driving. Many foundational works in computer vision benchmarked on KITTI. However, despite KITTI's success, it has fundamental limitations; namely, its small size and lack of full 360$^{\circ}$ camera coverage. Further, algorithms trained on KITTI have shown lower performance when transferred to other datasets~\cite{2020mmdet3d}, suggesting that training on KITTI may suffer from overfitting.

In recent years, major players including Waymo and Motional have released datasets more extensive than KITTI with multiple sensing modalities~\cite{2020waymodataset, 2020nuscenesdataset}. Despite this, many works still benchmark primarily on KITTI with only marginal improvements over prior results. To investigate, we scraped the KITTI leaderboard and selected all works from the top 50 places with a validated journal or conference publication. Of the 18 validated entries, many are recent: 13 were published in 2022; all have been released since 2020. They are all within 3\% on the leaderboard. Disappointingly, only a single entry ran experiments on KITTI, nuScenes~\cite{2020nuscenesdataset}, and Waymo~\cite{2020waymodataset} datasets, while one entry ran on KITTI and nuScenes, and eleven entries ran on KITTI and Waymo; see Appendix~\ref{app:percep-investigation} for the full table. 

\paragraph{Simulators.}
Simulators such as CARLA~\cite{2017carla} allow for important AV testing in a dynamic environment, provide closed-loop feedback (i.e.,~planning, control), and enable rare-event simulations that are difficult or dangerous to capture in the real world. However, CARLA comes with minimal resources to bootstrap AV DITA. In fact, CARLA provides no algorithms that do not use ground truth data nor do they provide an architecture for standing up one's own AV. As a result, there are few relevant AVs designs out of CARLA and few benchmark submissions to the CARLA challenge~\cite{2020carlaleadboard}. 

Recent works using CARLA have established success using end-to-end, vision-based reinforcement and imitation learning (RL, IL) in AVs (e.g.,~\cite{2017carla, 2021transfuser, 2020learningbycheating, 2020marlcarla}). The lean towards RL/IL is in part because CARLA has \emph{no built-in support for training or testing supervised learning algorithms}. There is no ``CARLA dataset'' and no clear way to generate one. There are several barriers to this, including that, to our knowledge, there is no way to obtain the list of objects in the field of view (even as a ground-truth oracle) without using a depth-sensor to determine if e.g.,~a building is blocking the view to the object. On the other hand, continuously running simulator trials with a reward function for RL/IL is easy. 

\paragraph{\textbf{Design Goal.}}

We designed \texttt{AVstack} to both bridge the dataset-simulator gap and to expand upon critically-absent features in existing APIs. We identified the core features \emph{essential} for wide dataset and simulator compatibility. The overarching theme of these innovations is simple: attach details to objects, not just documentation.

In particular, some features that enable \texttt{AVstack's} wide compatibility include: assigning attributes directly to sensor measurements so that each natively possesses all identifying information; expanding all object labels with 3D~bounding box, object type, object ID, velocity, acceleration, orientation, and angular velocity fields; defining flexible data structures to route sensor data to multiple end-points; standardizing reading, writing, and transforming sensor data and ground truth labels, and much more. Each feature is made possible by many precise decisions. For example, 3D~bounding boxes need clear reference frames (e.g.,~camera-frame, lidar-frame, ego-frame), orientation angle definitions (e.g.,~yaw=0 aligned with $x$-axis in camera frame, yaw-pitch-roll vs. roll-pitch-yaw ordering), bounding box height-offset (i.e.,~$0\coloneqq$~bottom of box, $0\coloneqq$~center~of~box).

Until now, each user 
would write software tailored towards the minimum required information for a single dataset -- the heterogeneity of options was too high a barrier to support multiple data sources. With \texttt{AVstack}, users can make software that is reusable and transferable between sources. Unlike Apollo, Autoware, and Pylot, \texttt{AVstack} is not designed towards a low-level, low-latency message-passing scheme. Rather, it exists in a high-level programming language quickly adaptable to new simulators and complex configurations. The reusability and adaptability lower the amount of effort required to test algorithms on different platforms. Now, we can start to expect more validation and verification of AV components on representative testing scenarios.

\subsubsection{Design Goal 1.2: Component Compatibility}
\paragraph{\textbf{Motivation.}}
A frustrating challenge to researchers is that groundbreaking AV components are difficult to use beyond their original benchmarks. To enable insightful evaluation of new components within a longitudinal environment, the developer must be able to quickly stand up and rearrange an AV using off-the-shelf components. While an individual benchmark may have many perception algorithms that can all be tested uniformly, minimal support exists to stitch that perception algorithm together in a longitudinal evaluation with tracking, motion prediction, and path planning.

Platforms that provide some degree of component compatibility, such as Pylot, do not support both datasets and simulators. Pylot also uses a custom message passing framework designed to minimize latency which is not useful for algorithm trade studies. Rather, it is suited for applications with real-time consideration.

\begin{figure}
    \centering
    \vspace{-4pt}
    \includegraphics[width=0.928\linewidth]{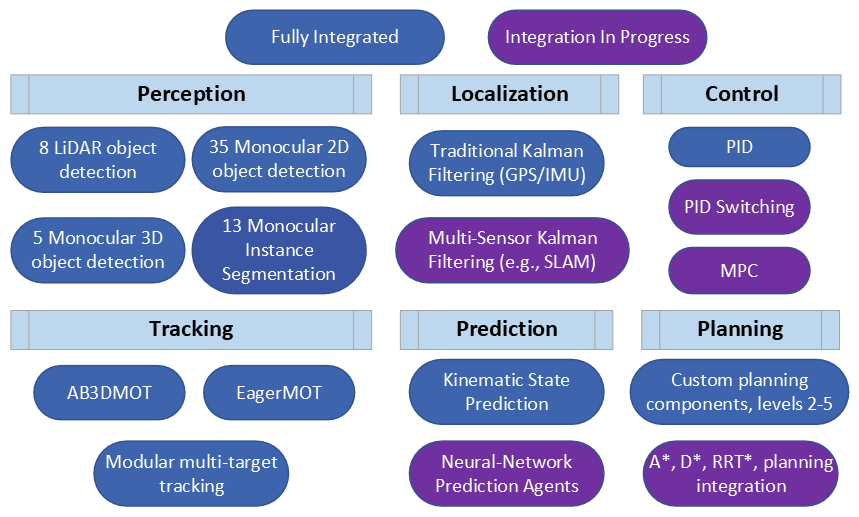}
    \vspace{-10pt}
    \caption{\texttt{AVstack} supports a broad set of open-source algorithms. Both custom implementations and open-source libraries bootstrap the diverse set of integrated capabilities. The set of compatible software is continuously growing.}
    \label{fig:avstack-algorithms}
\end{figure}

\paragraph{\textbf{Design Goal.}}
\texttt{AVstack} supports many prominent open-source components for both module-based design of AVs and end-to-end learning-based approaches. \texttt{AVstack} leans on existing open-source libraries to complement custom components. Fig.~\ref{fig:avstack-algorithms} provides a sample of the capabilities at the time of publication. In particular, \texttt{AVstack} currently supports 4 modes of perception with over 50 different perception algorithms. This wide compatibility is obtained under a common interface that allows for AV reconfiguration and reuse of algorithms and components.

\subsection{Design Goal 2: Unified Conventions}

There is no ``official'' set of conventions for AV datasets and simulators. This is for good reason: each dataset satisfies different needs of the end user. Datasets without positioning data may specify all objects in ego-relative coordinates (e.g.,~\cite{2013kittidataset}). Some datasets may introduce sensors not present in other datasets (e.g.,~radar in~\cite{2020nuscenesdataset}) while others defer all sensor specification to the user (e.g.,~\cite{2017carla}). Further, end users may have different state-vector requirements with degrees of freedom ranging from three (x, y, yaw; e.g.,~\cite{2000torcs}) to nine (x, y, z, height, width, length, roll, pitch, yaw; e.g.,~\cite{2017carla, 2020nuscenesdataset}).

In a review of state-of-the-art AV datasets and simulators, we find that no two sources share the same coordinate axes, frame origin, and orientation angle conventions. In fact, we find that even within single providers (e.g.,~\cite{2013kittidataset}) there can be discrepancies in the conventions. Similarities and differences are highlighted in Table~\ref{tab:source-differences}.

\begin{table*}[t]
  \centering
  \resizebox{\textwidth}{!}{
  
  \begin{tabular}{||c c c c c c c||} 
     \hline
     Source & Vehicle Frame & Ego Origin & Object Origin & Rotation & Sensors (\#, Rate) & KeyFrame Rate \\ [0.5ex] 
     \hline\hline
     KITTI Object   & RDF   & N/A           & Box Bottom & Euler (1D) & Camera (4, 10Hz), LiDAR (1, 10Hz) & 10Hz \\ 
     \hline
     KITTI Raw      & FLU   & N/A           & Box Bottom & Euler (1D) & Camera (4, 10Hz), LiDAR (1, 10Hz) & 10Hz\\
     \hline
     KITTI Odometry & FLU   & Camera 0      & N/A        & DCM        & Camera (4, 10Hz), LiDAR (1, 10Hz) & 10Hz\\
     \hline
     nuScenes       & FLU   & GP Rear Axle  & Box Center & Quaternion & \begin{tabular}{@{}c@{}}Camera (6, 12Hz), LiDAR (1, 20Hz), \\ Radar (5, 13Hz), GPS/IMU (1, 1000Hz)\end{tabular} & 2Hz \\ 
     \hline
     Waymo          & FLU   & Ego Center    & Box Center & Euler      & \begin{tabular}{@{}c@{}}Camera (5, 10Hz), Main LiDAR (1, 10Hz), \\ Peripheral LiDAR (4, 10Hz)\end{tabular} & 10Hz\\ 
     \hline
     CARLA          & FRU   & GP Ego Center  & Box Center & Euler (3D) & Many (user-specific) & N/A\\ 
     \hline
     TORCS          & FL(U) & BEV Ego Center & BEV Center & Euler (1D) & Many (user-specific) & N/A\\
     \hline\hline
     AVstack        & Any   & Any  & Any & Any & Any & Any\\
     \hline
    \end{tabular}
    }
    
    \caption{Minor differences in data design become major headaches for the developer. The ego reference and other objects can be specified with different coordinates, sensor origin, and rotation conventions. Each data source uses different sensors of varying rates and different attachments. AVstack handles transformations automatically. GP - ground projected; RDF - right, down, forward; FLU - forward, left, up; FRU - forward, right, up; DCM - direction cosine matrix; BEV - bird's eye view.}
    \label{tab:source-differences}
    \vspace{-18pt}
\end{table*}

\subsubsection{Design Goal 2.1: Reference Frames}

\begin{figure}[t]
    \centering
    \includegraphics[width=0.92\linewidth]{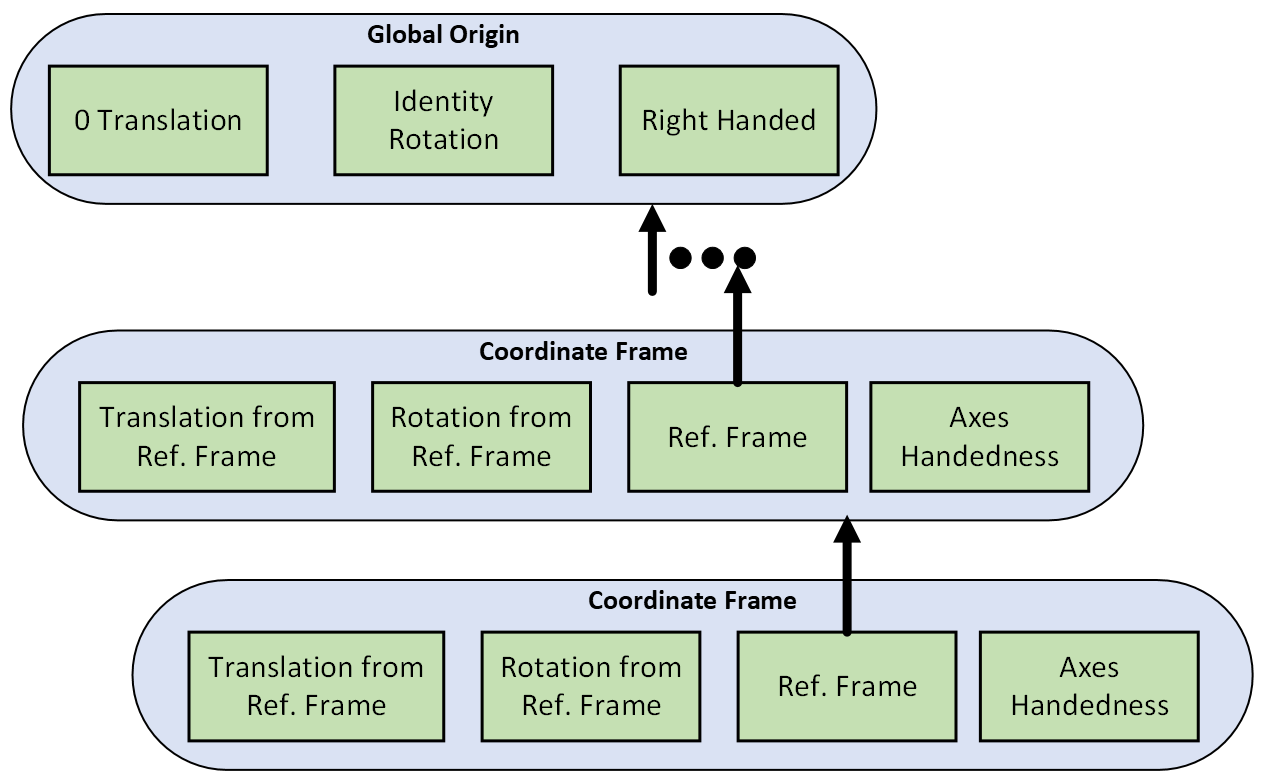}
    \vspace{-10pt}
    \caption{Reference frame defined by $(Tr,\, R,\, Ref\textrm{-}P,\, H)$, with $Tr$ a translation, $R$ a rotation, $Ref\textrm{-}P$ a parent reference, and $H$ the axes handedness. Reference frame chain of command (RefChoc) followed to a common ancestor when comparing objects or fusing data from complementary sensors/agents.}
    \label{fig:refchoc}
\end{figure}

\paragraph{\textbf{Motivation.}}
The level of complexity and lack of standard of reference frames hinders dataset-agnostic component design and introduces error into complex multi-sensor, multi-agent configurations. To mitigate this, we standardize reference frame definitions with a wrapper around each dataset and simulator. We also introduce the reference frame chain of command (RefChoC) that represents the dependency on secondary reference frames (see~Fig.~\ref{fig:refchoc}).

\begin{figure}[t]
    \centering
    \includegraphics[width=0.76\linewidth]{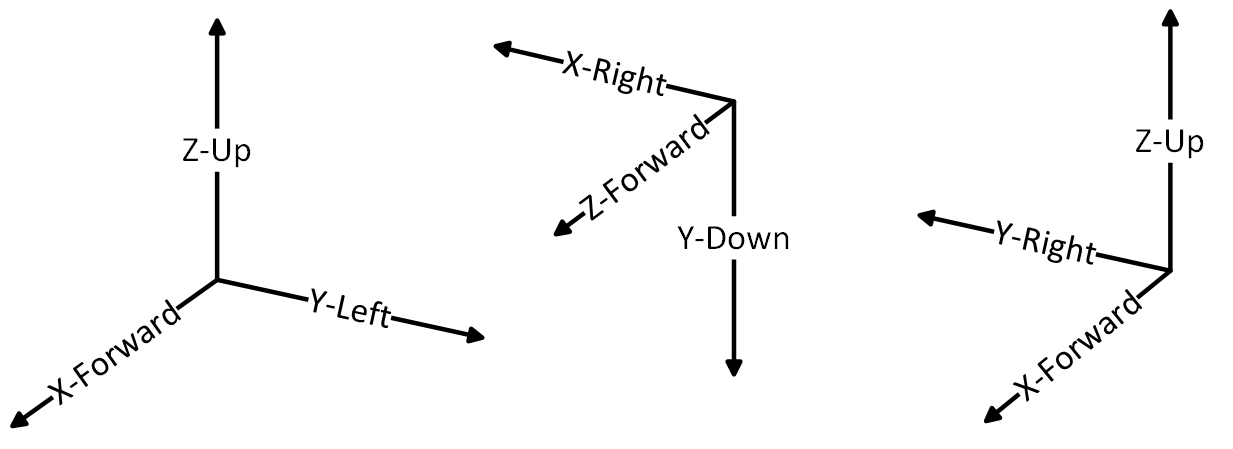}
    \vspace{-10pt}
    \caption{Sensors, datasets, and simulators often use different coordinate systems which may require up to a 4x4 transformation matrix to convert vectors between frames: (a) Standard frame employed by nuScenes~\cite{2020nuscenesdataset}, Waymo~\cite{2020waymodataset}; (b) Camera frame employed by KITTI~\cite{2013kittidataset}; (c) Left-handed frame employed by CARLA~\cite{2017carla}.}
    \label{fig:coordinates}
\end{figure}

\paragraph{Coordinate Axes.}
Coordinate frames cause headaches in even the most proficient of developers. This is particularly important when data sources use different conventions. One small difference is illustrated in Fig.~\ref{fig:coordinates}. While KITTI always labels objects in a right-down-forward (``camera'') coordinate frame, nuScenes and Waymo use frames dependent on the sensor which includes many possible orientations (see Appendix~\ref{app:dataset-design}). CARLA uses a non-traditional left-handed forward, right, up frame. 

\paragraph{Rotation Conventions.}
Many datasets represent orientation with Euler or Tait-Bryan (grouped under the name ``Euler'' in this work) angles (see Table~\ref{tab:source-differences}). This allows for a compact representation of the orientation that is (sometimes) human-interpretable. However, Euler angles are problematic for several reasons. The most obvious drawback is the lack of specificity: there are at least 12 accepted methods of specifying orientation using Euler conventions~\cite{1977angles}. This introduces error into the development process. Second, Euler angles suffer from gimbal lock and discontinuities in certain special cases. While this is seldom a problem in real-world driving, it is very relevant for AV simulators due to local nature of map coordinates.

\paragraph{Sensor Calibrations.}
Each sensor should be accompanied by a calibration that describes both where the sensor is positioned relative to the ego (often called: ``extrinsics'') and sensor-specific properties (often called: ``intrinsics''). Unfortunately, many datasets have ambiguous calibrations. KITTI provides calibration data but minimal instructions on how to use it or which data requires transformation. KITTI also only allows for ego-relative coordinates, which can impair target tracking models. Meanwhile, CARLA describes the unique conventions of its coordinate system but no supporting functions~in~the~software.

There is additional complexity beneath the surface across the board: calibrations must define \textit{whether the translation is in the pre-rotated or post-rotated reference frame}. 4~x~4 transformation matrices use post-rotation while it is most interpretable to use pre-rotation. Different providers take different approaches, and details are seldom documented. Furthermore, specifying a calibration is ambiguous, even under a clear reference frame and pre/post order if it does not specify which \textit{direction the transformation should be applied} (i.e.,~does it represent ``$A\rightarrow B$'' or ``$B\rightarrow A$''?).

\paragraph{\textbf{Design Goal.}}
We performed many iterations designing reference frames for the unified API of \texttt{AVstack}. To achieve standardized reference frames for the first time and provide a clear and elegant reference management solution, each physical object, bounding-box, sensor, and sensor measurement in \texttt{AVstack} is accompanied by a \texttt{calibration} and/or \texttt{origin} field. These are handled automatically by \texttt{AVstack} for the supported datasets (KITTI,~nuScenes,~CARLA).

Translations, vectors, rotations, and transformations are always relative to a reference coordinate frame. \texttt{AVstack} innovatively defines the reference frame as the tuple $Ref \coloneqq (Tr,\, R,\, Ref\textrm{-}P,\, H)$, with $Tr$ a translation, $R$ a rotation, $Ref\textrm{-}P$ a parent reference frame (for chained reference-frames, e.g.,~detection-to-sensor-to-ego-to-world), and $H$ the handedness of the axes. $(Tr,\, R)$ form the \texttt{origin} field. We illustrate \texttt{AVstack}'s approach for chained reference frames using a pass-by-reference approach in Fig.~\ref{fig:refchoc}. We call this approach the \emph{Reference Frame Chain of Command} (RefChoc). The RefChoc is the most reliable way to-date to support both simple cases of chaining (e.g.,~detection-to-sensor-to-ego-to-world) and complex cases (e.g.~multi-sensor, multi-agent) equally while implicitly handling coordinate transformations for the user to mitigate error-prone manual calculations.

\subsubsection{Design Goal 2.2: Relevant Metrics \& Evaluations}

\paragraph{\textbf{Motivation.}}
Metrics facilitate quantitative assessment of an autonomy stack's performance. Many popular self-driving and computer vision benchmarks~(e.g.,~\cite{2013kittidataset, 2020nuscenesdataset}) provide metrics at the component-level such as camera perception mean-average-precision (mAP), LiDAR perception mAP, tracking performance, prediction accuracy. These follow the hypothesis that improving individual components will lead to improved AVs in the aggregate.

The sum-of-its-parts argument neglects cross-cutting interactions and trade-offs that exist at the intersection between components. For example, many perception metrics neglect model runtime and the impact of latency on path planning and control. Similarly, improving individual components ignores inter-component error propagation; e.g.,~mAP takes the mean AP over all classes while not all classes impact motion prediction or path planning equally.

\paragraph{\textbf{Design Goal.}}
In response to the shortcomings of single-component metrics, we quantify performance at multiple components simultaneously, similar to~\cite{2021pylot}. \texttt{AVstack} provides a large selection of metrics at each level of the pipeline including the Responsibility Sensitive Safety (RSS) metric~\cite{2017rsssafety}. A select list of the supported metrics can be found in Table~\ref{tab:metrics}. Maintaining a broad set of metrics for longitudinal scenarios helps pursue:
\begin{table}[t!]
    \centering
    \begin{tabular}{||l l||}
        \hline
        \textbf{Module} & \textbf{Metric} \\
        \hline
        Perception & \tworowsubtableleft{False Positive Rate (FPR), Precision, mAP}{False Negative Rate (FNR), Recall, IoU}\\
        \hline
        Tracking   & \tworowsubtableleft{\tworowsubtableleft{IoU, False Track Rate (FTR)}{Missed Track Rate (MTR)~\cite{2008trackingeval}}}{\tworowsubtableleft{Higher Order Tracking Accuracy (HOTA)~\cite{2021hotametric}}{CLEAR~\cite{2008CLEARmetrics}, VACE~\cite{2017VACEmetrics}, IDEucl~\cite{2021hotametric}}} \\
        \hline
        Prediction & \tworowsubtableleft{Average Displacement Error (ADE)~\cite{2009ADEmetric}}{Final Displacement Error (FDE)~\cite{2016sociallstm}}\\
        \hline
        Planning   & \tworowsubtableleft{Responsibility Sensitive Safety (RSS)~\cite{2017rsssafety}}{Path KL Divergence~\cite{2020nuscenesdataset}}\\
        \hline
        Control    & \tworowsubtableleft{Responsibility Sensitive Safety (RSS)~\cite{2017rsssafety}}{CARLA Leaderboard Benchmark~\cite{2017carla}}\\
        \hline
    \end{tabular}
    \caption{\texttt{AVstack} unifies metrics for longitudinal testing while previous works only tested isolated components. \texttt{AVstack} uniquely incorporates the RSS safety metric.}
    \vspace{-20pt}
    \label{tab:metrics}
\end{table}

\begin{enumerate}
    \item \textbf{Cross-Cutting Interactions:} AV designers cannot ignore the interactions between components and the error propagation that exist when designing a longitudinal agent.
    \item \textbf{Longitudinal Analysis:} Single-frame examples from datasets are incapable of validating the full performance of AVs due to their complex temporal behavior.
    \item \textbf{Safety Evaluation:} Paradoxically, safety is both a primary method of regulating autonomy~\cite{2017cummingsreview} and woefully under-utilized in quantitatively evaluating AVs.
\end{enumerate}

\subsection{Design Goal 3: Modular Testing}

\texttt{AVstack} enables expanded AV lifecycle analysis. We describe how \texttt{AVstack}'s design enables for the first time reconfigurable architectures, expanded evaluations, streamlined model training, and multi-sensor, multi-agent configurations. 

\subsubsection{Design Goal 3.1: Reconfigurable Architectures}

\begin{figure}
    \centering
    \includegraphics[width=0.66\linewidth]{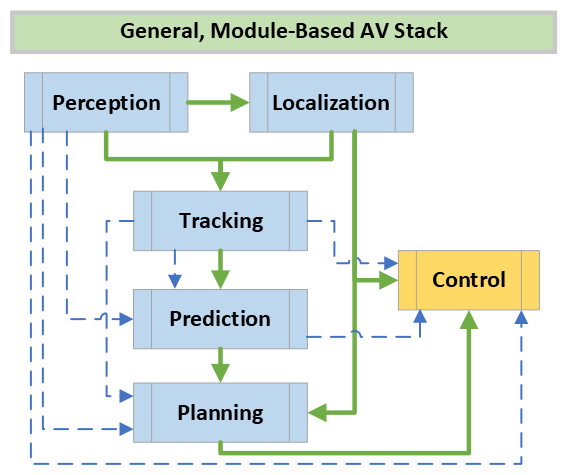}
    \vspace{-10pt}
    \caption{\texttt{AVstack} configuration is modular: any connection between modules is feasible. Breaking traditional constraints, any connection between \textit{components} is also feasible. Module names illustrate classic module-based AV design. Traditional connections in bold green; new connections in dashed blue.}
    \label{fig:modular-components}
\end{figure}

\paragraph{\textbf{Motivation.}}
Many open platforms constrain users to purely module-based~\cite{2021pylot} or purely end-to-end~\cite{2021transfuser}, which limits software reusability and next-level evaluations. Pylot, Apollo, and Autoware have rigid architectures (green lines in Fig.~\ref{fig:modular-components}) due to their low-level message passing. Changing architecture is difficult in all cases, and changing implementation in Apollo and Autoware is very challenging. It is more difficult to perform trade studies comparing sensors, to incorporate new sensors, and to consider novel AV architectures. These factors contribute to stagnation in AV development.

\paragraph{\textbf{Design Goal.}}
Components are the backbone of computation in AVs. In contrast to other platforms, \texttt{AVstack} enables \textit{any} connection between components with its reconfigurable design. The reconfigurable architecture cuts software complexity at the expense of real-time guarantees. Fig.~\ref{fig:modular-components} illustrates that \texttt{AVstack} opens up ``non-traditional'' connections between modules.

\begin{figure}
    \centering
    \includegraphics[width=0.962\linewidth]{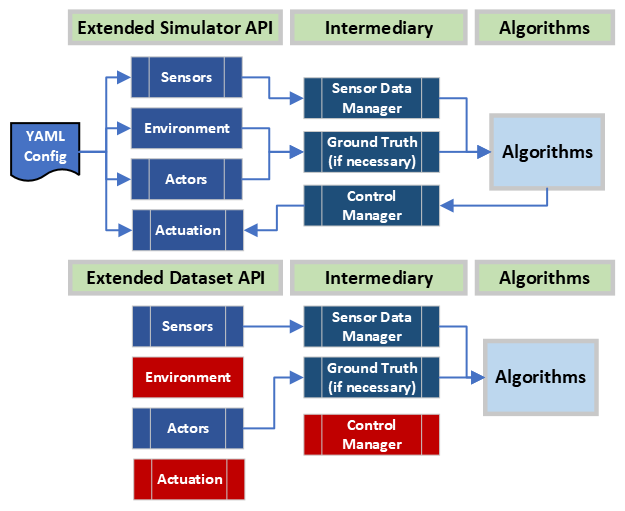}
    \vspace{-10pt}
    \caption{\texttt{AVstack} partitioned into API, intermediary, and algorithm modules to support reusability and portability of components. APIs built on common framework to allow first-of-a-kind dataset$\leftrightarrow$simulator transfer testing.}
    \label{fig:avstack-high-level}
\end{figure}

Imperatively, \texttt{AVstack}'s design philosophy disassociates implementation from platform. Thus, components are reusable between and among datasets and simulators. We illustrate in Fig.~\ref{fig:avstack-high-level} the flow of data. Simulator and dataset interfaces are standardized around base classes with common methods to get sensor data and object labels. The API is flexible enough to serve as the interface for all data sources. This supports early-stage development on captured datasets with longitudinal testing on end-to-end simulators.

\subsubsection{Design Goal 3.2: Expanded Evaluations}

\paragraph{\textbf{Motivation.}}
Our meta-analysis from Table~\ref{tab:kitti-publications} (Appendix~\ref{app:percep-investigation}) suggests that transfer testing of algorithms is too difficult with existing tools. Too few works perform testing on multiple large, complex datasets. Moreover, an even smaller set of works perform longitudinal analysis of inter-component error propagation. At the same time, simulators including CARLA do not provide sufficient resources to bootstrap AV implementations for longitudinal testing. 

\paragraph*{\textbf{Design Goal.}}
\texttt{AVstack} greatly expands evaluations for AVs. It enables dataset-to-dataset, dataset-to-simulator, and simulator-to-simulator transfer testing. AVs can be designed for static dataset, passive longitudinal, or active longitudinal (i.e.,~with control) self-driving scenarios. To show the deep level of insight made possible by \texttt{AVstack}, we present metrics from a large trade study across 5 AV configurations in Section~\ref{sec:4-use-cases-transfer}.

\subsubsection{Design Goal 3.3: Streamlined Learning}

\paragraph{\textbf{Motivation.}}
Supervised learning is a critical piece of AVs. Many modules including perception and path planning rely on learned components to perform fast and accurate inference on sensor data. A major challenge of learning-based techniques is that retraining is fraught with errors when trying to adapt datasets. Moreover, even mature simulations have limited ways to generate labeled training data from the simulator, even as a ground-truth oracle. There is no way to natively capture ground truth object labels in view of a sensor and unoccluded by buildings.

\paragraph{\textbf{Design Goal.}}
To aid the supervised learning process for AVs, we leverage mature modular infrastructures for supervised and reinforcement learning. \texttt{AVstack} uses MMLab's~\cite{2020mmdet3d} open-source training infrastructure and provides a custom \texttt{AVstack} dataset interface to train and test dozens of perception models. We also provide a methodology for generating training data from the CARLA simulator. \texttt{AVstack} implements much-needed automated methods for cleaning CARLA data such as field-of-view estimation, occlusion categorization, and bounding box projection to address critical barriers in the adoption of CARLA for realistic self-driving. In Section~\ref{sec:4-use-cases-collaborative}, we illustrate how this data generation process can be configured to generate complex multi-agent scenarios and collaborative V2V, V2I sensing data for model training and testing. This allows for creation of large volumes of collaborative perception data with consistent ground truth labels between multiple viewpoints.

\subsubsection{Design Goal 3.4: Multi-Sensor, Multi-Agent Systems}

\paragraph{\textbf{Motivation.}}
Multi-sensor and multi-agent testing are part of a critical wave of next-level challenges for AVs~\cite{2017cummingsreview}. As investments in smart infrastructure are considered, it is critical to evaluate the trade-offs in collaborative configurations. However, there are several barriers to testing both cases. Multi-sensor testing is difficult because sensor data always requires transformations between reference frames and may be configured with partially overlapping fields of view. Unfortunately, it is error-prone to leave multi-sensor configuration up to the developer; yet few public platforms provide effective multi-sensor support. Multi-agent testing has also yet to be sufficiently realized. The majority of evaluations in self-driving have focused on static datasets that lack multi-agent information. Similarly, even in simulator environments, mature AV research platforms have constrained architectures and components. This limited modularity means that adding new sensor data, integrating new components, and designing new algorithms is burdensome.

\begin{figure}
    \centering
    \includegraphics[width=0.94\linewidth]{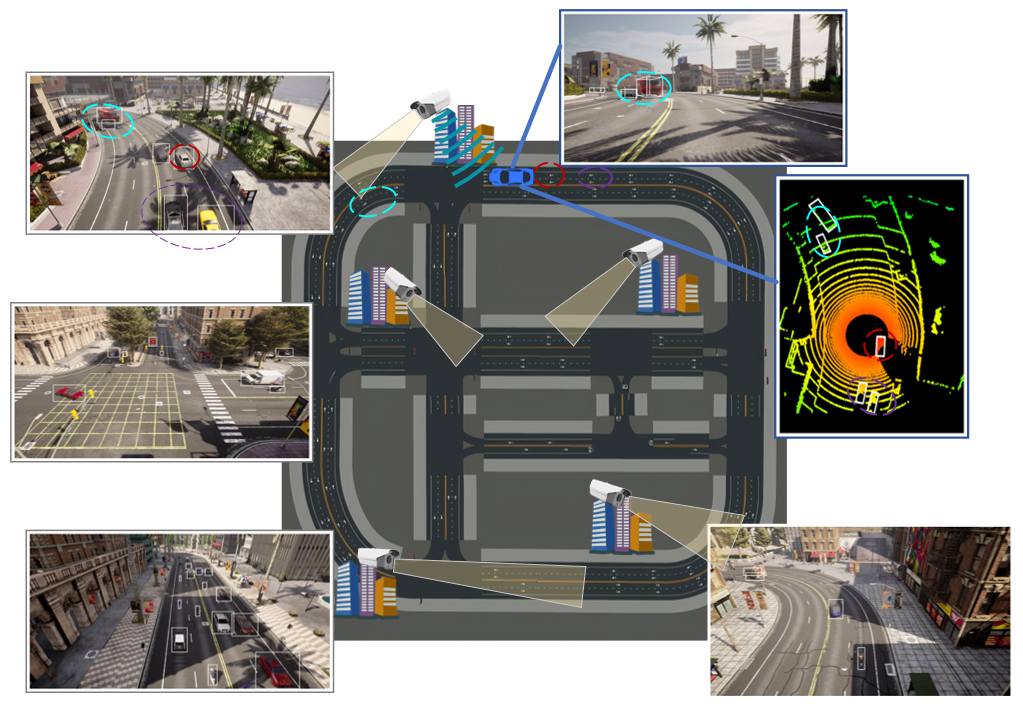}
    \vspace{-12pt}
    \caption{\texttt{AVstack} bootstraps CARLA for multi-sensor, multi-agent configurations, opening possibility for next-generation AV DITA. Here, camera sensors are placed at infrastructure locations to provide collaborative view. (top right) Ego vehicle with local camera and LiDAR: object detections in white boxes. (top left) Infrastructure sensor detects objects in camera corresponding to objects from ego: cyan, red,~purple~ovals.}
    \label{fig:avstack-collaborative}
\end{figure}

\paragraph{\textbf{Design Goal.}}
To solve the sensor data and reference-frame challenges in multi-sensor/multi-agent configurations, \texttt{AVstack} has several important innovations. First, reference frame transformations can be performed automatically by specifying a start and end-point reference. This removes error-prone coordinate transformations (e.g., \emph{object-to-sensor1-to-ego-to-sensor2} for multi-sensor; \emph{sensor1-to-agent1-to-world-to-agent2-to-sensor2} for multi-agent). Second, \texttt{AVstack} has a growing list of sensors to which it offers compatibility. In the simulator context, \texttt{AVstack} bootstraps ego and sensor classes with clearer and developer-friendly configurations to support existing simulator features. Third, architecture design is modular in \texttt{AVstack}. Components from a single-agent AV can be reused in multi-agent contexts. Single-agent AVs can be evaluated against multi-agent AVs in a unified simulation framework (Fig.~\ref{fig:avstack-collaborative}).

\subsection{Design Goal 4: Low Barrier}
\paragraph{\textbf{Motivation.}}
Apollo, Autoware, and Pylot are mature AV platforms but all have a high barrier to entry. All rely on high-performance message passing frameworks to deliver low-latency sensor data at the cost of architecture flexibility. Source code for Apollo and Autoware is complex and rigid. They are targeted to full-stack AVs that ingest sensor data and output control decisions. This makes debugging individual algorithms and components incredibly difficult; changing AV architecture is exceptionally challenging. 

\paragraph{\textbf{Design Goal.}}
\texttt{AVstack} provides a low-barrier and flexible AV testing framework. For the first time, there is compatibility between datasets and simulators. At the intermediary between data and algorithms are thread-safe data structures that handle flexible routing of data from source to destination in a high-level programming language. Our no-copy philosophy allows data to be transferred efficiently to support near-real-time execution; however, data are handled with the utmost flexibility for the user. An object-oriented approach allows sensor data to be efficiently routed with multiple end-points. In Section~\ref{sec:4-use-cases}, we provide case studies using just dozens of lines of code on top of \texttt{AVstack} to create unique AVs and diverse testing environments.

\section{Use Case Experiments} \label{sec:4-use-cases}

In this section, we show how \texttt{AVstack} enables important exploration, trade studies, and analysis at low development cost.

\subsection{Portability and Transfer Testing} \label{sec:4-use-cases-transfer}
Two major causes of a slowdown in AV development are poor infrastructures for transfer testing between datasets \& simulators, and limited longitudinal evaluations. The ability to perform algorithm testing across data sources is vital for validation of complex components. Running longitudinal evaluations helps understand cross-component error propagation, which is lacking in single-component analysis.

\begin{figure}
\centering
    \includegraphics[width=0.924\linewidth]{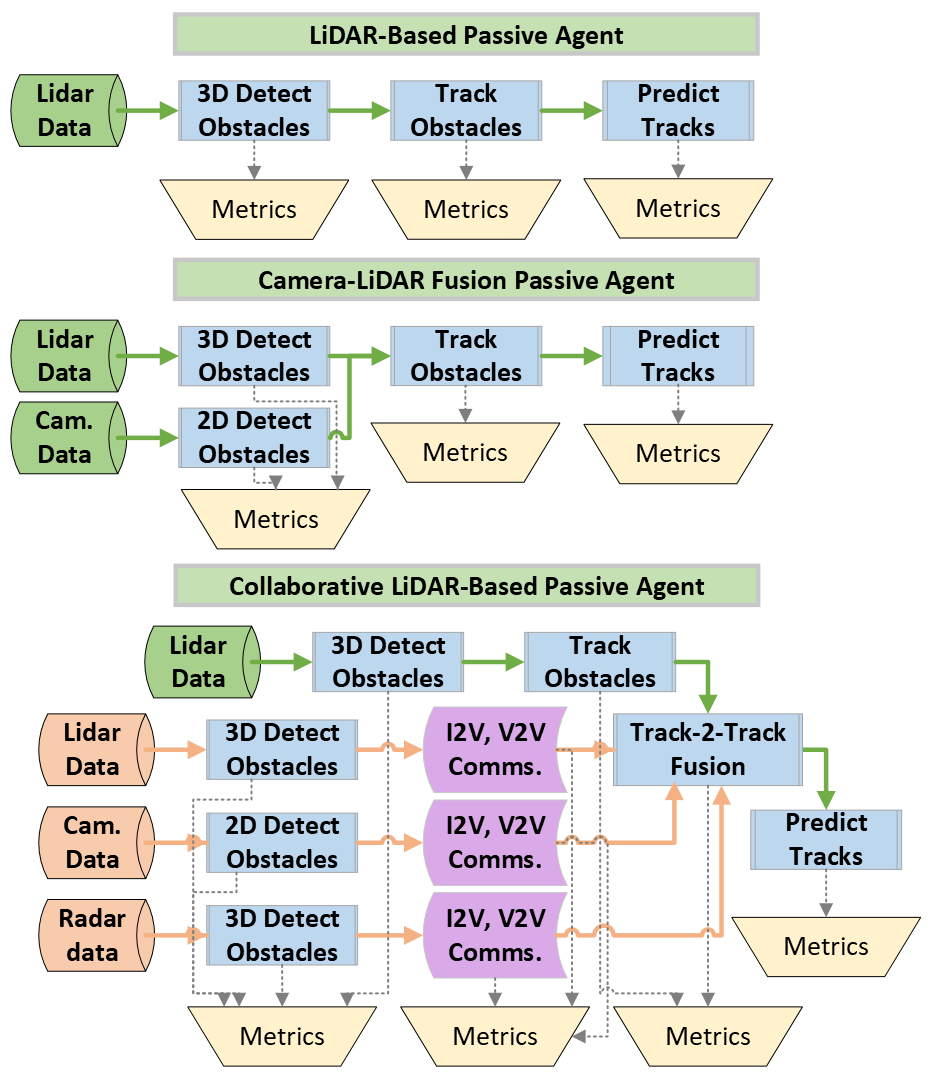}
    \vspace{-10pt}
    \caption{LiDAR-only, camera-LiDAR, and collaborative LiDAR agents require only 15, 20, and 30 lines of code at the top level to instantiate. Implementations can run on datasets \& simulators allowing for insightful and rapid trade studies.}
    \label{fig:passive-agents}
\end{figure}

To demonstrate that \texttt{AVstack} enables transferability between data sources, we design passive agents using LiDAR-based and camera-LiDAR fusion component architectures (e.g., as in~\cite{hallyburton_security22,hallyburton2023securing}) shown in Fig.~\ref{fig:passive-agents}. We can use \texttt{AVstack} to create these dataset-agnostic agents using just 15 and 20 line of code. We call these ``passive'' because we leave out planning and control -- a capability made possible by \texttt{AVstack}'s reconfigurable design. Within the two architectures, we test different combinations of algorithms to form five different implementations. The complete case study specification is represented with a ``configuration table'' in \texttt{AVstack}, as illustrated in Table~\ref{tab:passive-study}. With this configuration table, \texttt{AVstack} evaluates the different AV implementations over KITTI, nuScenes, and CARLA on 10 randomly sampled longitudinal sequences. During each run, \texttt{AVstack} captures per-frame and per-sequence metrics that were summarized in Table~\ref{tab:metrics}.

\begin{table}[t]
    \centering
    \resizebox{\columnwidth}{!}{%
    \begin{tabular}{|c|l|l|l|l|}
        \hline
        \textbf{ID} & \textbf{LiDAR Percep} & \textbf{Cam Percep} & \textbf{Tracking} & \textbf{Prediction} \\
        \hline
        0 & PointPillars~\cite{2019pointpillars} & N/A & AB3DMOT~\cite{2020ab3dmot} & Kinematic \\
        \hline
        1 & 3DSSD~\cite{20203dssd} & N/A & AB3DMOT~\cite{2020ab3dmot} & Kinematic \\
        \hline
        2 & PointPillars~\cite{2019pointpillars} & FasterRCNN~\cite{2015fasterrcnn} & EagerMOT~\cite{2021eagermot} & Kinematic \\
        \hline
        3 & 3DSSD~\cite{20203dssd} & FasterRCNN~\cite{2015fasterrcnn} & EagerMOT~\cite{2021eagermot} & Kinematic \\
        \hline
        4 & PointPillars~\cite{2019pointpillars} & \tworowsubtableleft{Cascade}{MaskRCNN~\cite{2017maskrcnn}} & EagerMOT~\cite{2021eagermot} & Kinematic \\
        \hline\hline
    \end{tabular}}
    \caption{\texttt{AVstack} enables transferability between data sources. uses configuration tables to run trade studies. Modules are widely compatible with community implementations. Components are dataset-agnostic. The 5 case studies here are used for the transfer test in Section~\ref{sec:4-use-cases-transfer}.}
    \vspace{-16pt}
    \label{tab:passive-study}
\end{table}

\texttt{AVstack}'s output of the trade study is a set of detailed per-frame and per-case results (not shown) and an aggregated benchmark table; see Table~\ref{tab:passive-agent-results}. Videos of select sequences can be found online~\cite{avstack}. \texttt{AVstack}'s breadth and depth of measurements make it useful for component-wise analysis of AVs. In this case study, we find 3D object precision is high across all algorithms and all datasets; however, recall on nuScenes is low. Similarly, nuScenes tracking performance (HOTA) is lower compared to KITTI and CARLA. 
\begin{table*}[t]
  \centering
  \resizebox{\textwidth}{!}{%

\begin{tabular}{rllllllllll}
\toprule
 Case &                                                  Data &                                                                                         Per: 3D Prec. &                                                                                          Per: 3D Rec. &                                                                                         Per: 2D Prec. &                                                                                          Per: 2D Rec. &                                                                                             Trk: HOTA &                                                                                              Trk: MOTA &                                                                                             Trk: MOTP &                                                                                             Pred: ADE &                                                                                              Pred: FDE \\ \midrule
\midrule
    0 & \tworowsubtablecenter{K}{\tworowsubtablecenter{N}{C}} & \tworowsubtablecenter{0.37 +/- 0.24}{\tworowsubtablecenter{0.99 +/- 0.01}{\underline{0.99 +/- 0.00}}} & \tworowsubtablecenter{\underline{0.90 +/- 0.21}}{\tworowsubtablecenter{0.25 +/- 0.07}{0.77 +/- 0.02}} &                                           \tworowsubtablecenter{N/A}{\tworowsubtablecenter{N/A}{N/A}} &                                           \tworowsubtablecenter{N/A}{\tworowsubtablecenter{N/A}{N/A}} & \tworowsubtablecenter{\underline{0.52 +/- 0.19}}{\tworowsubtablecenter{0.11 +/- 0.04}{0.51 +/- 0.07}} & \tworowsubtablecenter{-0.03 +/- 2.48}{\tworowsubtablecenter{0.20 +/- 0.07}{\underline{0.48 +/- 0.05}}} & \tworowsubtablecenter{\underline{3.62 +/- 0.10}}{\tworowsubtablecenter{2.68 +/- 0.16}{2.85 +/- 0.14}} & \tworowsubtablecenter{1.33 +/- 0.94}{\tworowsubtablecenter{\underline{0.26 +/- 0.08}}{4.99 +/- 3.65}} & \tworowsubtablecenter{3.87 +/- 1.41}{\tworowsubtablecenter{\underline{0.26 +/- 0.08}}{11.55 +/- 5.37}} \\ \midrule
    1 & \tworowsubtablecenter{K}{\tworowsubtablecenter{N}{C}} & \tworowsubtablecenter{0.25 +/- 0.19}{\tworowsubtablecenter{\underline{1.00 +/- 0.02}}{0.99 +/- 0.00}} & \tworowsubtablecenter{0.39 +/- 0.17}{\tworowsubtablecenter{0.19 +/- 0.06}{\underline{0.68 +/- 0.05}}} &                                           \tworowsubtablecenter{N/A}{\tworowsubtablecenter{N/A}{N/A}} &                                           \tworowsubtablecenter{N/A}{\tworowsubtablecenter{N/A}{N/A}} & \tworowsubtablecenter{0.40 +/- 0.16}{\tworowsubtablecenter{0.09 +/- 0.04}{\underline{0.46 +/- 0.07}}} & \tworowsubtablecenter{-0.15 +/- 0.32}{\tworowsubtablecenter{0.12 +/- 0.05}{\underline{0.43 +/- 0.04}}} & \tworowsubtablecenter{\underline{3.07 +/- 0.73}}{\tworowsubtablecenter{2.75 +/- 0.13}{2.82 +/- 0.13}} & \tworowsubtablecenter{1.20 +/- 0.78}{\tworowsubtablecenter{\underline{0.31 +/- 0.06}}{5.20 +/- 3.33}} & \tworowsubtablecenter{1.86 +/- 1.77}{\tworowsubtablecenter{\underline{0.31 +/- 0.06}}{12.11 +/- 4.91}} \\ \midrule
    2 & \tworowsubtablecenter{K}{\tworowsubtablecenter{N}{C}} & \tworowsubtablecenter{0.37 +/- 0.24}{\tworowsubtablecenter{\underline{0.69 +/- 0.18}}{0.62 +/- 0.13}} & \tworowsubtablecenter{\underline{0.90 +/- 0.21}}{\tworowsubtablecenter{0.32 +/- 0.02}{0.88 +/- 0.06}} & \tworowsubtablecenter{0.31 +/- 0.18}{\tworowsubtablecenter{\underline{0.90 +/- 0.04}}{0.40 +/- 0.25}} & \tworowsubtablecenter{\underline{0.73 +/- 0.20}}{\tworowsubtablecenter{0.52 +/- 0.11}{0.13 +/- 0.09}} & \tworowsubtablecenter{\underline{0.71 +/- 0.13}}{\tworowsubtablecenter{0.11 +/- 0.04}{0.12 +/- 0.05}} &  \tworowsubtablecenter{\underline{0.60 +/- 0.22}}{\tworowsubtablecenter{0.11 +/- 0.07}{0.08 +/- 0.05}} & \tworowsubtablecenter{\underline{3.77 +/- 0.17}}{\tworowsubtablecenter{2.89 +/- 0.30}{3.00 +/- 0.30}} & \tworowsubtablecenter{\underline{0.77 +/- 0.58}}{\tworowsubtablecenter{1.05 +/- 0.43}{1.26 +/- 0.54}} &  \tworowsubtablecenter{1.78 +/- 1.67}{\tworowsubtablecenter{\underline{1.05 +/- 0.43}}{3.56 +/- 0.70}} \\ \midrule
    3 & \tworowsubtablecenter{K}{\tworowsubtablecenter{N}{C}} & \tworowsubtablecenter{0.25 +/- 0.19}{\tworowsubtablecenter{\underline{0.67 +/- 0.19}}{0.54 +/- 0.13}} & \tworowsubtablecenter{0.39 +/- 0.17}{\tworowsubtablecenter{0.24 +/- 0.03}{\underline{0.69 +/- 0.09}}} & \tworowsubtablecenter{0.31 +/- 0.18}{\tworowsubtablecenter{\underline{0.90 +/- 0.04}}{0.40 +/- 0.25}} & \tworowsubtablecenter{\underline{0.73 +/- 0.20}}{\tworowsubtablecenter{0.52 +/- 0.11}{0.13 +/- 0.09}} & \tworowsubtablecenter{\underline{0.46 +/- 0.18}}{\tworowsubtablecenter{0.09 +/- 0.03}{0.10 +/- 0.05}} &  \tworowsubtablecenter{\underline{0.34 +/- 0.14}}{\tworowsubtablecenter{0.07 +/- 0.03}{0.06 +/- 0.05}} & \tworowsubtablecenter{2.98 +/- 0.55}{\tworowsubtablecenter{2.93 +/- 0.29}{\underline{2.99 +/- 0.28}}} & \tworowsubtablecenter{\underline{0.61 +/- 0.69}}{\tworowsubtablecenter{0.63 +/- 0.59}{1.44 +/- 0.41}} &  \tworowsubtablecenter{1.21 +/- 1.95}{\tworowsubtablecenter{\underline{0.63 +/- 0.59}}{3.38 +/- 0.54}} \\ \midrule
    4 & \tworowsubtablecenter{K}{\tworowsubtablecenter{N}{C}} & \tworowsubtablecenter{0.37 +/- 0.24}{\tworowsubtablecenter{\underline{0.69 +/- 0.18}}{0.62 +/- 0.13}} & \tworowsubtablecenter{\underline{0.90 +/- 0.21}}{\tworowsubtablecenter{0.32 +/- 0.02}{0.88 +/- 0.06}} & \tworowsubtablecenter{0.29 +/- 0.18}{\tworowsubtablecenter{0.78 +/- 0.05}{\underline{0.93 +/- 0.02}}} & \tworowsubtablecenter{\underline{0.95 +/- 0.23}}{\tworowsubtablecenter{0.72 +/- 0.08}{0.60 +/- 0.07}} & \tworowsubtablecenter{\underline{0.70 +/- 0.08}}{\tworowsubtablecenter{0.12 +/- 0.03}{0.36 +/- 0.09}} &  \tworowsubtablecenter{\underline{0.59 +/- 0.08}}{\tworowsubtablecenter{0.10 +/- 0.08}{0.30 +/- 0.07}} & \tworowsubtablecenter{\underline{3.77 +/- 0.12}}{\tworowsubtablecenter{2.89 +/- 0.10}{3.01 +/- 0.07}} & \tworowsubtablecenter{\underline{0.86 +/- 0.23}}{\tworowsubtablecenter{1.06 +/- 0.33}{2.68 +/- 1.32}} &  \tworowsubtablecenter{1.51 +/- 1.12}{\tworowsubtablecenter{\underline{1.06 +/- 0.33}}{5.17 +/- 3.80}} \\ \midrule
\bottomrule
\end{tabular}
}
  \caption{\texttt{AVstack} enables first-of-a-kind trade studies simply by specifying a configuration table such as Table~\ref{tab:passive-study}. Results are averaged over 10 longitudinal trials using the centrall mounted LiDAR and forward-facing camera. Each trial is over a 20 second scene for each dataset (K: KITTI, N: nuScenes, C: CARLA). Each AV configuration ``Case" is described in Table~\ref{tab:passive-study}. For the first time, metrics can be computed at each level of the pipeline (Per. 2D/3D: 2D or 3D Perception, Trk.: Tracking, Pred.: Prediction) at the same time to illuminate error propagation between modules. Best performance is highlighted per-cell.}
  \label{tab:passive-agent-results}
    \vspace{-10pt}
\end{table*}





\subsection{Multi-Sensor, Multi-Agent Collaboration} \label{sec:4-use-cases-collaborative}

While multi-sensor, multi-agent configurations are imperative for next-generation AV evaluations, they are difficult to design and test using today's available platforms. Some recent works have begun to analyze cooperative settings using ad-hoc development environments~\cite{2022cooperative, 2022openv2v}. Previous evaluation platforms have leveraged existing datasets to run experiments. Usefully, \texttt{AVstack} is not tied to an individual dataset. Rather, the \texttt{AVstack} API provides a flexible and easy to use approach to leverage existing datasets and to generate \textit{any} scenario, including multi-sensor, multi-agent configurations, in the CARLA simulator.

We use \texttt{AVstack} to design a collaborative agent with an architecture similar to Fig.~\ref{fig:passive-agents}(c). The agent possesses a LiDAR sensor with a limited range of $25~m$. To obtain sufficient situational awareness, the agent must use information from nearby infrastructure sensors to complement its own limited sensing information. We do not consider planning or control components and instead investigate the agent just using perception, tracking, and prediction performance. 

We use the \texttt{AVstack} API to test our multi-agent design. We place 40 64-line LiDARs with a field-of-view of $180^\circ$ at random locations in CARLA's Town-10. These serve as the infrastructure sensors. Each collaborative sensor is placed at a $30^\circ$ pitch angle and a height of 15~m to obtain an appropriate viewing angle. We chose to use LiDAR sensors to simplify 3D positioning, but any and all sensors in CARLA can be used, including cameras and radars. 

With this configuration, we design two trade study experiments to evaluate the trade-offs between (1) sensor communication range and detection accuracy, and (2) sensor rate and detection accuracy. Table~\ref{tab:use-case-collaborative} highlights the different configurations in this experiment.

Using the trade study capability of \texttt{AVstack}, we run the ego agent over the 9 collaborative cases from Table~\ref{tab:use-case-collaborative} on 5 randomly-generated CARLA scenes with 150 ``other'' vehicles. Collaborative detections are transmitted from sensor to agent at the specified data rate. Upon receiving messages, the agent first performs preprocessing to ignore any detections outside of a 100~m radius, for computational efficiency. The agent then integrates detections with data association, assigns measurements to existing tracked objects, and spawns new tracks with unassigned detections. Additional configuration details can be found at~\cite{avstack} as well as in Appendix~\ref{app:collab-case}.

At the culmination of the study, AVstack generates aggregated results tables, shown in Table~\ref{tab:collaborative-agent-results}. Videos of select sequences can be found online~\cite{avstack}. We find that collaborative sensing can aid an agent, particularly in this case where the ego's sensor range was limited. In Table~\ref{tab:collaborative-agent-results}-A, we find the HOTA metric is highest (best) for C1-Ideal and C1-1. Also, prediction error, ADE and FDE, are lower with collaboration compared to C1-base. In Table~\ref{tab:collaborative-agent-results}-B, we find that tracking performance does not deteriorate when trading sensor rate from $10~Hz$ to $5~Hz$ for a decrease in detection noise - HOTA remains constant among all cases. While differences in prediction errors, ADE and FDE, are not significantly different between cases C2-1 and C2-2, it is worth investigating in more detail the impact of sensor rate on prediction performance.

\begin{table}[t]
    \centering
    \caption{Collaborative Experiment Design}
    \vspace{-10pt}
    \label{tab:use-case-collaborative}
    \resizebox{\columnwidth}{!}{%
    \begin{tabular}{|c|l|l|l|l|}
        \hline
        \textbf{Case} & \textbf{LiDAR Percep} & \textbf{Cam Percep} & \textbf{Tracking} & \textbf{Prediction} \\
        \hline \hline
        All & PointPillars~\cite{2019pointpillars} & N/A & AB3DMOT~\cite{2020ab3dmot} & Kinematic \\
        \hline
    \end{tabular}}
    \subcaption*{Panel A: AV configuration constant for all collaborative studies.}
  \vspace{-10pt}
    \bigskip
    \resizebox{\columnwidth}{!}{%
    \begin{tabular}{|c|l|l|l|l|l|}
        \hline
        \textbf{Case} & \textbf{Density} & \textbf{Det. Type} & \textbf{Comm Range} & \textbf{Det. Rate} & \textbf{Det. Noise} \\
        \hline \hline
        C1-Ideal & 40/map & 3D Box & 100~m & 10~Hz & None\\
        C1-1 & 40/map & 3D Box     & 100~m & 10~Hz & High\\
        C1-2 & 40/map & 3D Box     & 70~m  & 10~Hz & Med\\
        C1-3 & 40/map & 3D Box     & 50~m  & 10~Hz & Low\\
        C1-Base & 40/map & N/A & N/A & N/A & N/A\\
        \hline
    \end{tabular}}
    \subcaption*{Panel B: Experiment (C1) trading comm range for noise.}
  \vspace{-10pt}
    \bigskip
    \resizebox{\columnwidth}{!}{%
    \begin{tabular}{|c|l|l|l|l|l|}
        \hline
        \textbf{Case} & \textbf{Density} & \textbf{Sensor} & \textbf{Comm Range} & \textbf{Det. Rate} & \textbf{Det. Noise} \\
        \hline \hline
        C2-Ideal & 40/map & 3D Box & 80~m & 10~Hz & None\\
        C2-1 & 40/map & 3D Box     & 80~m  & 10~Hz  & High\\
        C2-2 & 40/map & 3D Box     & 80~m  & 5~Hz  & Low\\
        C2-Base & 40/map & N/A & N/A & N/A & N/A\\
        \hline
    \end{tabular}}
    \subcaption*{Panel C: Experiment (C2) tests communication rate vs. noise.}
      \vspace{-10pt}
\end{table}

\begin{table*}[t]
  \centering
  \caption{Collaborative Vehicle-to-Infrastructure Case Study Results.}
  \vspace{-10pt}
  \label{tab:collaborative-agent-results}
    \resizebox{\textwidth}{!}{%
\begin{tabular}{rllllllll}
\toprule
 Case & Data & Collab: Sensors-in-range/frame & Collab: Dets/frame &     Trk: HOTA &      Trk: MOTA &     Trk: MOTP &     Pred: ADE &     Pred: FDE \\ \midrule
\midrule
    C1-Ideal &    C &             13.00 +/- 3.30 &   122.00 +/- 63.69 & 0.55 +/- 0.16 & -0.40 +/- 0.61 & 3.10 +/- 0.08 & 1.19 +/- 0.53 & 3.23 +/- 1.86 \\ \midrule
    C1-1 &    C &             13.00 +/- 3.30 &   122.00 +/- 63.69 & 0.52 +/- 0.14 & -0.46 +/- 0.64 & 2.94 +/- 0.06 & 0.86 +/- 0.29 & 2.65 +/- 1.23 \\ \midrule
    C1-2 &    C &              5.00 +/- 2.45 &    55.00 +/- 35.72 & 0.32 +/- 0.14 & -0.90 +/- 1.55 & 2.92 +/- 0.09 & 1.41 +/- 0.14 & 3.64 +/- 0.46 \\ \midrule
    C1-3 &    C &              2.00 +/- 1.89 &    20.00 +/- 41.96 & 0.54 +/- 0.12 & -0.24 +/- 0.47 & 2.99 +/- 0.05 & 1.07 +/- 0.24 & 2.65 +/- 0.75 \\ \midrule
    C1-Base &    C &                        N/A &                N/A & 0.47 +/- 0.09 &  0.35 +/- 0.07 & 3.10 +/- 0.10 & 2.42 +/- 1.99 & 6.40 +/- 3.26 \\ \midrule
\bottomrule
\end{tabular}
    }
    \subcaption*{Panel A: Trading communication range for detection accuracy over 10 trials of 500 frames in CARLA.} \label{tab:collaborative-agent-results-A}
\vspace{-4pt}
    \bigskip
    \resizebox{\textwidth}{!}{%
    \begin{tabular}{rllllllll}
    \toprule
     Case & Data & Collab: \#S-in-range/frame & Collab: Dets/frame &     Trk: HOTA &      Trk: MOTA &     Trk: MOTP &     Pred: ADE &     Pred: FDE \\ \midrule
    \midrule
        C2-Ideal &    C &              3.50 +/- 1.50 &    40.25 +/- 13.75 & 0.63 +/- 0.17 & -0.06 +/- 0.68 & 3.04 +/- 0.07 & 1.36 +/- 0.45 & 3.36 +/- 1.31 \\ \midrule
        C2-1 &    C &              3.50 +/- 1.50 &    40.25 +/- 13.75 & 0.61 +/- 0.18 & -0.08 +/- 0.69 & 2.90 +/- 0.07 & 0.98 +/- 0.14 & 2.23 +/- 0.60 \\ \midrule
        C2-2 &    C &              3.50 +/- 1.50 &     18.50 +/- 6.00 & 0.60 +/- 0.18 & -0.09 +/- 0.67 & 2.88 +/- 0.06 & 0.94 +/- 0.01 & 2.05 +/- 0.13 \\ \midrule
        C2-Base &    C &                        N/A &                N/A & 0.66 +/- 0.08 &  0.54 +/- 0.07 & 3.02 +/- 0.10 & 1.60 +/- 0.56 & 5.74 +/- 0.74 \\ \midrule
    \bottomrule
    \end{tabular}
    }
    \subcaption*{Panel B: Trading communication rate for detection accuracy over 10 trials of 500 frames in CARLA.} \label{tab:tab:collaborative-agent-results-B}
    \vspace{-10pt}
\end{table*}

\vspace{-4pt}
\section{Conclusion} \label{sec:conclusion}
\vspace{-4pt}

We have introduced \texttt{AVstack} as an open-source, reconfigurable software platform for AV design, implementation, test, and analysis. We have illustrated in several case studies that \texttt{AVstack} supports rapid prototyping of reusable AV components, longitudinal evaluations with component-wise metrics, and diverse multi-sensor, multi-agent configurations. \texttt{AVstack} delivers solutions to the most common challenges faced by AV users with its bank of community-support components, by bridging convention conflicts among datasets and simulators, by supporting algorithm reuse with dataset-agnostic and flexible components, and by delivering much-needed support for next-level analysis. Its key design principles will help accelerate the push toward important AV milestones. In several case studies focusing on portability and transfer testing, as well as testing of multi-sensor, multi-agent collaboration, we have illustrated these benefits of the use of \texttt{AVStack}.

\begin{acks}
This work is sponsored in part by the ONR under the agreements N00014-20-1-2745 and N00014-23-1-2206, AFOSR award number FA9550-19-1-0169, NSF CNS-1652544 award as well as the National AI Institute for Edge Computing Leveraging Next Generation Wireless Networks, Grant CNS-2112562.
\end{acks}

\vspace{-12pt}
\bibliographystyle{ieeetr}
\bibliography{references_pruned}

\appendix
\section{Intentional Design Omissions} \label{app:design-omissions}
No platform can satisfy the requirements of all AV use-cases because some are in conflict. For example, introducing architecture modularity can sacrifice real-time performance. To address some of the critical barriers to AV development, \textit{a modular research platform is essential and lacking}. 

We are faced with fundamental architecture questions for multi-sensor, multi-agent AVs where industry is dramatically outpacing research. For the next generation of smart vehicles, insightful DITA must be prioritized. To do so in an expeditious manner, there must be a low barrier to entry, even if this means sacrificing other qualities. In particular, \texttt{AVstack} intentionally places less emphasis on the following areas:
\begin{itemize}
    \item \textbf{Real Time:} \texttt{AVstack} is not proposed as a real-time solution. We have not performed experiments evaluating latency. Attempting to package \texttt{AVstack} as a real-time AV may require a real-time operating system and low-latency data passing which would negatively affect modularity.
    \item \textbf{Low-Level Programming:} \texttt{AVstack} is based on Python to allow for rapid prototyping and easy interfacing to third-party simulation engines. It was not written with speed or memory as a primary goal, in contrast to higher-barrier autonomy stacks Apollo and Autoware.
\end{itemize}

\begin{table}[!t]
    \centering
    \resizebox{\columnwidth}{!}{%
    \begin{tabular}{|c c l l l|}
        \hline\hline
         Friendly Name & Year & KITTI & nuScenes & Waymo  \\
         \hline
         Sparse Fuse Dense & 2022 & Y (84.8) & N & N \\
          \hline
         CasA & 2022 & Y (84.0) & N & Y (78.3/69.6) \\
          \hline
         GLENet & 2022 & Y (83.2) & N & Y (77.3/69.7) \\
          \hline
         VPFNet & 2022 & Y (83.2) & N & N \\
          \hline
         Graph R-CNN & 2022 & Y (83.2) & N & Y (72.6/72.1) \\
          \hline
         BtcDet & 2022 & Y (82.9) & N & Y (78.6/70.1) \\
          \hline
         SPG & 2021 & Y (82.7) & N & Y \\
          \hline
         SE-SSD & 2021 & Y (82.5) & N & N \\
          \hline
         DVF & 2022 & Y (82.5) & N & Y (67.6/62.7)\\
          \hline
         RDIoU & 2022 & Y (82.3) & N & Y (78.4/69.5) \\
          \hline
         FocalsConv & 2022 & Y (82.3) & Y (70.1) & Y (72.2/64.1)\\
          \hline
         CLOCs & 2020 & Y (82.3) & N & N\\
       \hline
         SASA & 2022 & Y (82.2) & Y (45) & N \\
       \hline
         VoTr & 2021 & Y (82.1) & N & Y (69.0/60.2)\\
       \hline
         Pyramid R-CNN & 2021 & Y (82.1) & N & (76.3/67.0)\\
       \hline
         VoxSet & 2022 & Y (82.1) & N & Y (77.9/70.2)\\
       \hline
         SRIF-RCNN & 2022 & Y (82.0) & N & N \\
       \hline
         Q-Net & 2022 & Y (82.0) & N & N\\
         \hline \hline
    \end{tabular}}
    \caption{Recent publications atop the KITTI leaderboard are not always cross-validated against other, larger datasets. The nuScenes dataset has limited adoption. Continued testing on KITTI has only achieved marginal improvements on already high performing marks.}
    \vspace{-16pt}
    \label{tab:kitti-publications}
\end{table}

\section{State of the Art Perception} \label{app:percep-investigation}
The KITTI dataset~\cite{2013kittidataset} was instrumental in the progress of AV perception development. Since, KITTI's original release, major players including Waymo and Motional have released datasets more extensive than KITTI with multiple sensing modalities~\cite{2020waymodataset, 2020nuscenesdataset}. Unfortunately, we find that even recent state-of-the-art perception algorithms neglect to provide sufficient evaluation on these more challenging datasets. To investigate, we scraped perception benchmark leaderboards, as described in Section~\ref{sec:3-design-1.1-datasets_sims}. The findings of this meta-analysis are in Table~\ref{tab:kitti-publications}. Of 18 validated entries in the top 50 on KITTI, many are recent, and progress between them has been marginal at only 3\% gained. Unfortunately, even these recent works neglect cross-dataset evaluations, leading to challenges with reproducibility and translational success in contexts such as simulators and real AVs.



\section{KITTI, nuScenes, Waymo Configurations} \label{app:dataset-design}
The release of high-fidelity benchmark datasets from major research institutions and prominent industry players has significantly contributed to a boom in AV algorithm development. Large datasets like nuScenes~\cite{2020nuscenesdataset} and Waymo's Open Dataset~\cite{2020waymodataset} have garnered attention recently for their challenging mix of weather conditions and multiple complementary sensing modalities.

\begin{figure}[!t]
    \centering
    \begin{subfigure}[b]{0.68\linewidth}
        \includegraphics[width=\textwidth]{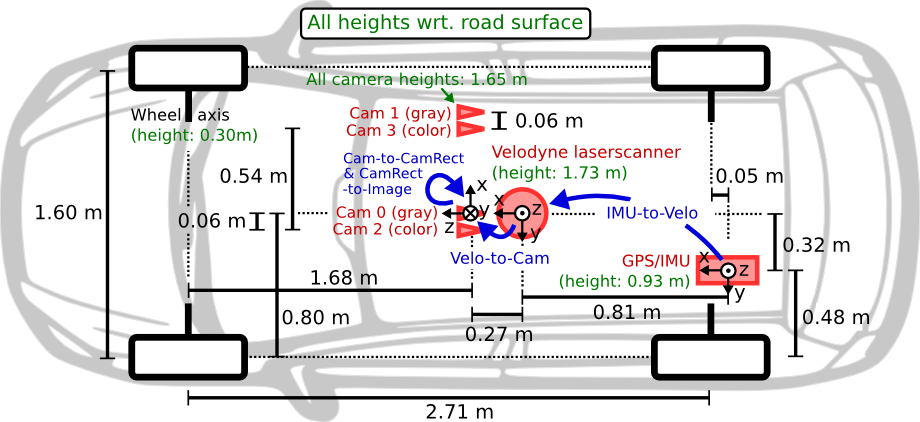}
        \subcaption{KITTI~\cite{2013kittidataset}}
    \end{subfigure}
    \begin{subfigure}[b]{0.74\linewidth}
        \includegraphics[width=\textwidth]{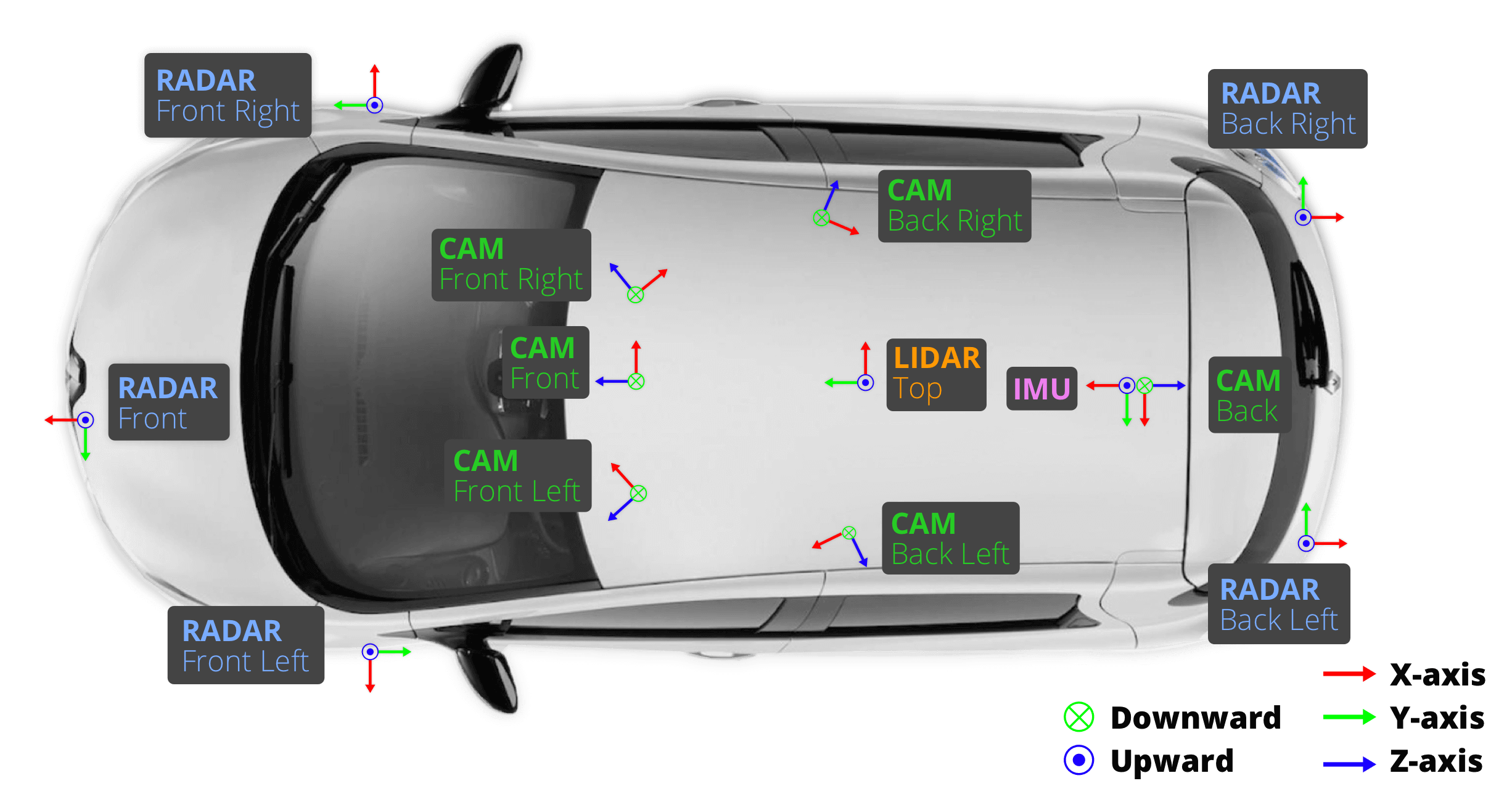}
        \subcaption{nuScenes~\cite{2020nuscenesdataset}}
    \end{subfigure}
    \begin{subfigure}[b]{\linewidth}
        \includegraphics[width=\textwidth]{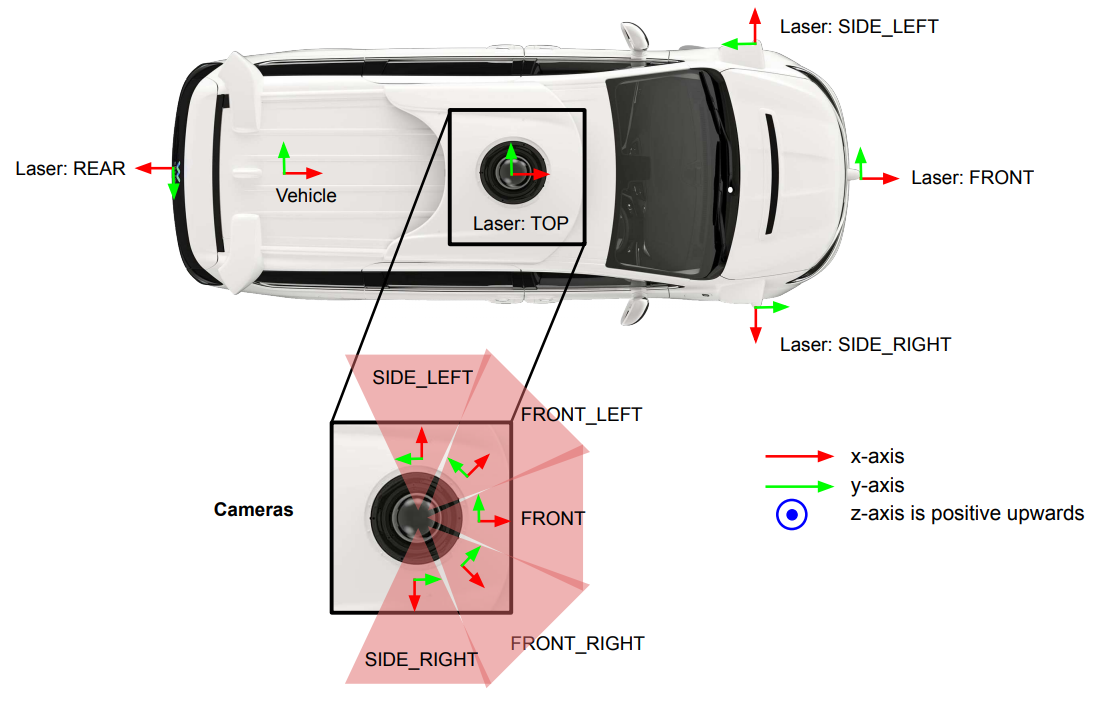}
        \subcaption{Waymo~\cite{2020waymodataset}}
    \end{subfigure}
    \vspace{-20pt}
    \caption{Configurations from different AV data source providers are all unique; each dataset has its own sensor types, sensor orientations, reference frames, and data rates. Evaluating components across all sources leads to insightful~results.}
    \label{fig:dataset-cars-and-sensors}
\end{figure}

Despite their contributions to the field, no platform has managed to unify the datasets under an tractable umbrella. This is in part due to the intricacy and uniqueness of each platform itself. To help illuminate why unifying these datasets under a common interface is challenging, we provide the sensor configurations for KITTI~\cite{2013kittidataset}, nuScenes, and Waymo's open dataset in Figure~\ref{fig:dataset-cars-and-sensors}.

\section{Configuration of Collaborative Case Study} \label{app:collab-case}
The vehicle-to-infrastructure (V2I) collaborative case study of Section~\ref{sec:4-use-cases-collaborative} provides a framework for future efforts to develop multi-agent components and design smart cities. In this section, we provide additional details on the specific parameters used. These details can also be found in the source code online at~\cite{avstack}. We used LiDAR sensors as our infrastructure sensors. In pre-processing, we determined if objects were in the field of view of a sensor for ground-truth evaluation by using ray-tracing to filter out objects that were completely occluded (i.e.,~no LiDAR points in bounding box). We did so because CARLA has no alternative method, to our knowledge, of validating if an object is in view of a sensor. After pre-processing, we simulated detections from the LiDAR sensor rather than run a perception algorithm. This was solely so that we could apply our own noise model to the infrastructure detections as a trade study. Then, to simulate V2I communication, we performed range-based filtering to identify which infrastructure sensors were in-range of the ego vehicle. Detections were passed with no latency to the ego agent. The agent then fused the infrastructure detections with existing tracks in a Kalman filter with a standard assignment algorithm. We evaluated performance of the ego agent against objects in the field of view of the ego within a range of 100~m and a maximum occlusion score of ``partial''.

\end{document}